\documentclass[journal,twoside,web]{ieeecolor}
\usepackage{tmi}
\usepackage{cite}
\let\labelindent\relax
\usepackage{enumitem}
\usepackage{threeparttable}
\usepackage{amsmath,amssymb,amsfonts}
\usepackage{graphicx}
\usepackage{multirow, makecell}
\usepackage{bm}
\usepackage{algorithm}
\usepackage{algorithmic}
\usepackage{colortbl}
\usepackage{arydshln}
\usepackage{multirow}
\usepackage{multicol}
\usepackage{color}
\usepackage{float}
\usepackage{url}
\usepackage{times}
\usepackage{epsfig}
\usepackage{mathtools}
\usepackage{multirow}
\usepackage{diagbox}
\usepackage{booktabs}
\usepackage[misc]{ifsym}
\usepackage{authblk}
\usepackage{textcomp}

\newcommand{\ie}{\textit{i}.\textit{e}.}
\newcommand{\eg}{\textit{e}.\textit{g}.}
\newcommand{\tabincell}[2]{\begin{tabular}{@{}#1@{}}#2\end{tabular}}
\def\BibTeX{{\rm B\kern-.05em{\sc i\kern-.025em b}\kern-.08em
    T\kern-.1667em\lower.7ex\hbox{E}\kern-.125emX}}
\begin{document}
\title{Boundary-aware Supervoxel-level Iteratively Refined Interactive 3D Image Segmentation with Multi-agent Reinforcement Learning}
\author{Chaofan Ma, Qisen Xu, Xiangfeng Wang, Bo Jin, Xiaoyun Zhang, Yanfeng Wang, and Ya Zhang

\thanks{Manuscript received November 6, 2020; revised December 16, 2020; accepted December 27, 2020. Date of publication December 31, 2020; date of current version September 30, 2021. This work was supported in part by the National Key Research and Development Program of China under Grant 2020AAA0107400, in part by NSFC under Grant 12071145, in part by the Shanghai Municipal Commission of Economy and Informatization (SHEITC) under Grant 2018-RGZN-02046, and in part by the Science and Technology Commission of Shanghai Municipality (STCSM) under Grant 18DZ2270700 and Grant 20511101100. (Corresponding author: Ya Zhang.) }
\thanks{Chaofan Ma, Xiaoyun Zhang, Yanfeng Wang, and Ya Zhang are with the Cooperative Medianet Innovation Center, Shanghai Jiao Tong University, Shanghai 200240, China (e-mail: chaofanma@sjtu.edu.cn; xiaoyun.zhang@sjtu.edu.cn; wangyanfeng@sjtu.edu.cn; ya\_zhang@sjtu.edu.cn). }
\thanks{Qisen Xu, Xiangfeng Wang, and Bo Jin are with the MOE Key Laboratory for Advanced Theory and Application in Statistics and Data Science, School of Computer Science and Technology, East China Normal University, Shanghai 200062, China, and also with the Shanghai Institute of Intelligent Science and Technology, Tongji University, Shanghai 200092, China (e-mail: 51184501067@stu.ecnu.edu.cn; xfwang@cs.ecnu.edu.cn; bjin@cs.ecnu.edu.cn).}
\thanks{Digital Object Identifier 10.1109/TMI.2020.3048477}
}

\maketitle

\begin{abstract}

Interactive segmentation has recently been explored to effectively and efficiently harvest high-quality segmentation masks by iteratively incorporating user hints. While iterative in nature, most existing interactive segmentation methods tend to ignore the dynamics of successive interactions and take each interaction independently. 
We here propose to model iterative interactive image segmentation with a Markov decision process (MDP) and solve it with reinforcement learning (RL) where each voxel is treated as an agent. 
Considering the large exploration space for voxel-wise prediction and the dependence among neighboring voxels for the segmentation tasks, multi-agent reinforcement learning is adopted, where the voxel-level policy is shared among agents.
Considering that boundary voxels are more important for segmentation, we further introduce a boundary-aware reward, which consists of a global reward in the form of relative cross-entropy gain, to update the policy in a constrained direction, and a boundary reward in the form of relative weight, to emphasize the correctness of boundary predictions. To combine the advantages of different types of interactions, \ie, simple and efficient for point-clicking, and stable and robust for scribbles, we propose a supervoxel-clicking based interaction design. 
Experimental results on four  benchmark datasets have shown that the proposed method significantly outperforms the state-of-the-arts, with the advantage of fewer interactions, higher accuracy, and enhanced robustness.

\end{abstract}


\section{Introduction}
\label{sec:introduction}
\IEEEPARstart{M}{edical} image segmentation is fundamental and critically important for a broad spectrum of medical imaging tasks such as measurement of organs and lesions, diagnosis and treatment planning. A variety of deep learning methods have been proposed for automatic medical image segmentation and shown great promise~\cite{milletari2016v,ronneberger2015u,zhou2018unet++}. 
However, a bottleneck for their wide application lies in requiring a large volume of manually labeled data for training which is labor-intensive and hard to meet for many application scenarios. 
As a result, the accuracy and robustness of the current automatic methods often fail to meet the standard for clinical use. 
Interactive segmentation is introduced to leverage user interactions, mostly in the form of clicks, scribbles, or bounding boxes, to obtain accurate segmentation through iteratively refinement~\cite{bredell2018iterative,cciccek20163d,wang2018deepigeos}.
While typically iterative in nature, the current interactive segmentation methods usually treat each refinement process independently and fail to capture the dynamics of successive interactions. 
As a result, they are unable to ensure that the corrected segmentation always stays. In other words, one error prediction can be corrected with user interactions and mispredicted again in a later iteration, which may result in stagnation or even degradation in performance.

In this paper, as an extension of our previous work~\cite{liao2020iteratively}, we model iteratively refined interactive image segmentation problem as a Markov decision process (MDP) and use reinforcement learning to learn a policy that maps states to probabilities of taking each available primitive action to update the segmentation results. In this way, at each refinement iteration, the model updates the labels of all voxels by taking consideration of not only the current state but also the previous state and its actions.
To reduce the exploration space to a tractable size and at the same time explicitly model the dependencies among voxels, the multi-agent reinforcement learning (MARL) is employed, where all agents are required to share the same policy and thus collaborate and communicate with each other through convolutional layers. 

The design of reward is critical for the success of  reinforcement learning. Considering that boundary voxels are more important for segmentation, we here propose a \emph{boundary-aware reward}, consisting of the \emph{global reward} and the \emph{boundary reward}. The global reward measures the relative cross-entropy gain between two consecutive iterations, with a positive reward for improved segmentation so that the new prediction is forced to outperform the previous one. The boundary reward emphasizes the correctness of the segmentation boundary by measuring the distance between each voxel and the ground truth boundary voxels. While the global reward ensures refinement convergence, the boundary reward is expected to further improve the accuracy of segmentation.

Another important factor for interactive segmentation is the form of user interactions. Point-clicking is fast and easy, but is susceptible and less accurate. Scribbles and bounding boxes are more stable and robust, but demand more efforts of users. To combine the advantage of the above two types of interactions, we propose a \emph{supervoxel-clicking} based interaction design, \ie, clicking on a point suggests that the corresponding supervoxel is error prone. The supervoxel-clicking based interaction is expected to be more robust to random variations in clicking positions.

Integrating the above design of reward and interaction to the MARL-based framework, we here propose a novel \textit{Boundary-aware Supervoxel-level Iteratively Refined Interactive Segmentation} (BS-IRIS) for 3D medical images. 
While most interactive segmentation methods require an initial segmentation mask (e.g. obtained with a certain automatic CNN)~\cite{liao2020iteratively,wang2018deepigeos, bredell2018iterative}, BS-IRIS deals with the setting where users label from scratch. 
We evaluate BS-IRIS with four publicly available challenge datasets, BraTS2015~\cite{menze:hal-00935640}, BraTS2019~\cite{bakas2018identifying}, MM-WHS~\cite{zhuang2016multi}, and NCI-ISBI2013~\cite{bloch2015nci}, which covers medical images of brain tumor, heart, and prostate. 
The experimental results have demonstrated that BS-IRIS  outperforms not only the state-of-the-art automatic CNNs in each corresponding challenges, but also several state-of-the-art interactive medical image segmentation methods. 
Furthermore, BS-IRIS is shown to be robust to different interaction quality, faster to converge, and requires less human interactions.

This paper is an extension of our preliminary work~\cite{liao2020iteratively}.
Modeling the interactive segmentation task with an MDP, both methods adopt a MARL-based framework for interactive segmentation of 3D medical images. 
The contributions of this paper on top of~\cite{liao2020iteratively} are summarized as follows.
\begin{itemize}[leftmargin=12pt,itemsep=0pt,parsep=0pt]
    \item A novel boundary-aware reward function that dynamically incorporates the boundary information and encourages agents to explore more efficiently.
    \item A new form of supervoxel-clicking based interaction with the combined advantage of point clicking and scribbles.
\end{itemize}

\section{Related work}
\label{sec:related work}
Interactive image segmentation has been widely applied to both natural~\cite{boykov2001interactive,xu2016deep} and medical images~\cite{rajchl2016deepcut,wang2018deepigeos}, where annotators provide certain hints to segmentation models to achieve better results. This section briefly reviews deep reinforcement learning and interactive image segmentation.

\subsection{Deep Reinforcement Learning}
Deep reinforcement learning (DRL) are getting increasing attention due to its promising performance. 
Deep Q-Network (DQN)~\cite{mnih2015human} is the first to combine deep learning with reinforcement learning. 
Generally speaking, reinforcement learning algorithms are categorized into value-based  and policy-based. 
The value-based algorithms, such as vanilla DQN~\cite{mnih2015human} and its varients (\eg, double DQN\cite{hasselt2016deep}, dueling DQN\cite{wang2016dueling} and prioritized experience replay~\cite{schaul2016prioritized}, etc.), utilize a Q-function to estimate the reward expectation of the action performed in the current state, and execute the policy according to this Q-function. 
Different from the value-based algorithms, policy-based algorithms, such as Reinforce~\cite{sutton1999policy}, directly model the policy. 
Actor-Critic based algorithms, such as A3C~\cite{mnih2016asynchronous}, DDPG~\cite{lillicrap2015continuous}, TRPO~\cite{schulman2015trust} and
PPO~\cite{schulman2017proximal}, combines the advantages of these two types of algorithms, which enables the resulted reinforcement learning algorithms to be more stable and achieve better performance.

\subsection{Traditional Interactive Image Segmentation}

Traditional methods make use of low-level features such as the histogram and similarities between pixels. GraphCut~\cite{boykov2001interactive} and GrabCut~\cite{rother2004grabcut} incorporate user hints into Max-Flow Min-Cut algorithm~\cite{boykov2004experimental}. 
DenseCRF~\cite{krahenbuhl2011efficient} considers pixel relations from neighbors to all pixel pairs. Criminisi \textit{et al.}~\cite{criminisi2008geos} proposes to use geodesic distance to calculate the distance between pixels, which is sensitive to contrast and suitable for medical images. Slic-Seg~\cite{wang2016slic} introduces a segmentation method for fetal MRI by learning from user annotations in only one slice.

\subsection{CNN-based Interactive Image Segmentation}

Many CNN-based algorithms have been recently developed for interactive image segmentation tasks.
iFCN~\cite{xu2016deep} is the first to propose an interactive object segmentation algorithm based on CNN. 
RIS-Net~\cite{liew2017regional} combines the segmentation of a full image and improvement of the local area. 
Li \textit{et al.}~\cite{li2018interactive} explores the multi-modality segmentation result space and solves the fuzziness caused by user clicking.
DEXTR~\cite{maninis2018deep} reduces user annotations from the bounding box to four extreme points with the same amount of user hints information.
Le \textit{et al.}~\cite{le2018interactive} includes an encoder-decoder network that takes both the image and user interactions.
In~\cite{hu2019fully}, the image and user interaction are divided into two networks, and the multi-scale fusion can let user hints having a more direct impact on the result.
Kontogianni \textit{et al.}~\cite{kontogianni2019continuous} consider user corrections as sparse training examples to update the model.
Agustsson \textit{et al.}~\cite{agustsson2019interactive} use shared scribble across regions, and allow the annotator to focus on the largest errors across the whole image.
Jang \textit{et al.}~\cite{jang2019interactive} and Sofiiuk \textit{et al.}~\cite{sofiiuk2020f} develop the backpropagating refinement scheme (BRS), which corrects the mislabeled pixels.
FCA-Net~\cite{lin2020interactive} treats all interaction points discriminately, and considers the difference between the first click and the remaining ones.
IOG~\cite{zhang2020interactive} leverages an inside point that is clicked at the center and two outside points at the  corner locations to encloses the target object.
DeepIGeoS~\cite{wang2018deepigeos} uses one CNN to obtain an initial coarse segmentation, and another CNN to take the user interactions as input and refine the initial coarse segmentation.
Inter-CNN~\cite{bredell2018iterative} extends DeepIGeoS to an iterative version, which iteratively refines the previous refined binary prediction. 
All the above algorithms ignore  the dynamic process for successive interactions. 

\begin{figure*}[ht]
\centering
\includegraphics[width=0.73\textwidth]{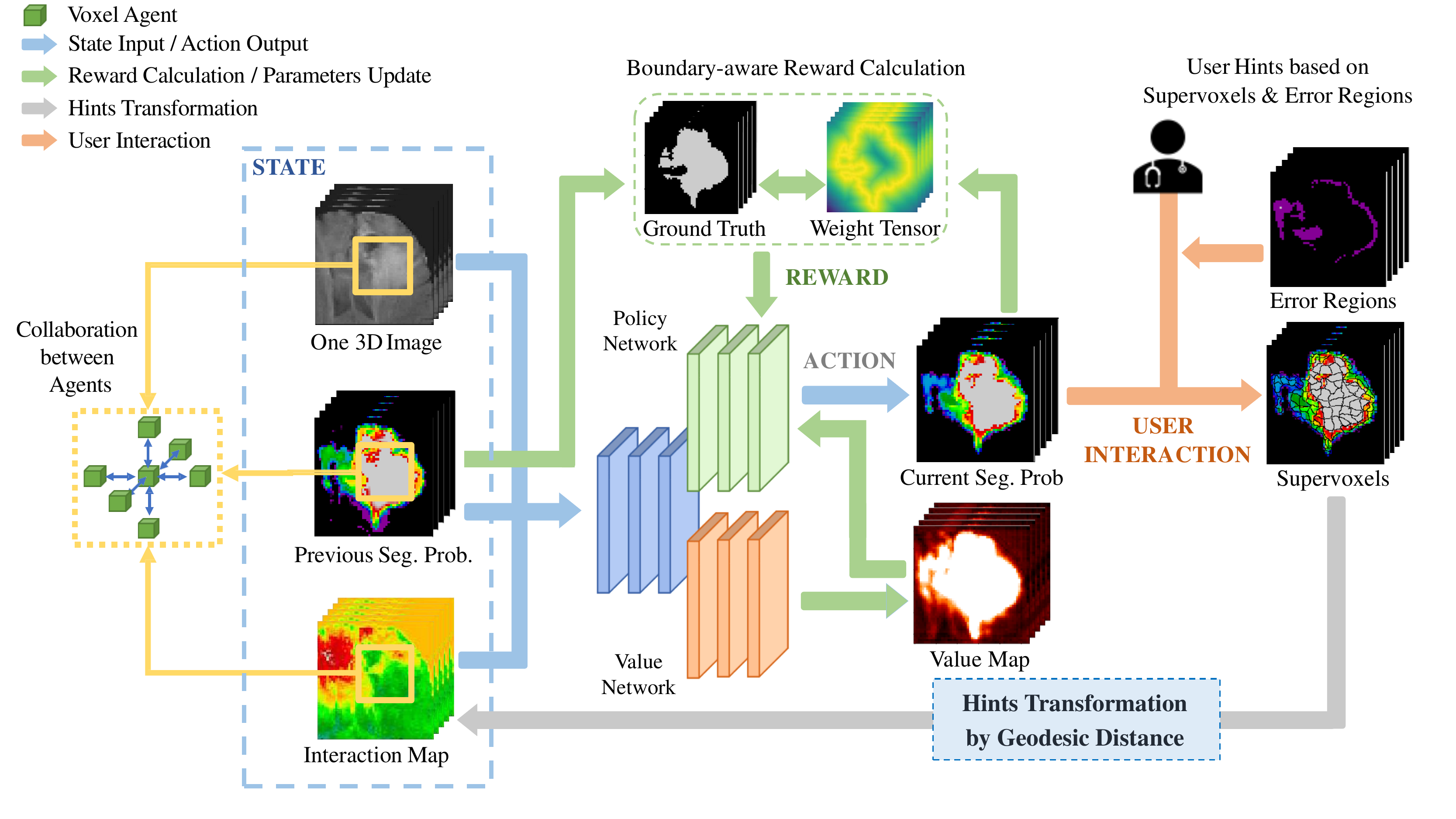}
\caption{The overview of BS-IRIS.}
\label{framework}
\end{figure*}
\subsection{RL-based Interactive Image Segmentation}
Some recent methods leverage RL to explicitly model the dynamics in interactive image segmentation tasks.
IAD~\cite{uijlings2018learning} uses RL for bounding box annotation.
SeedNet~\cite{song2018seednet} simulates user behaviors using reinforcement learning. It inputs the generated clicks into a split model, and then segment automatically in interactive form. Different from SeedNet, we investigate how reinforcement learning can be used by an interactive segmentation model.
Polygon-RNN~\cite{castrejon2017annotating} casts segmentation as a polygon prediction task, which takes an image crop as input and sequentially produces vertices of the polygon outlining the object. Human annotators only need to correct a vertex if needed. 
Based on this work, Polygon-RNN++~\cite{acuna2018efficient} designs a new CNN encoder architecture using RL and graph neural networks. 
However, the polygon-based methods cannot be applied to our tasks due to the incompatibility of 3D images with polygon segmentation, and the extremely large action space even with the meshing strategy.

\section{Methodology}
\label{sec:Methodology}

We formulate the dynamic process of iterative refined interactive image segmentation as an MDP. 
Following PixelRL~\cite{furuta2019fully} and IteR-MRL~\cite{liao2020iteratively}, we treat each voxel in a 3D image as a collaborative agent and employ multi-agent reinforcement learning (MARL) to solve this problem.
The framework of BS-IRIS is shown in Fig. \ref{framework}.
At each refinement iteration, the state information contains an image, previous segmentation, and the interaction.
Given the current state, a policy network outputs a set of actions, \ie, different scale of probability adjustments, and a value network estimates the values of the current state, \ie, value map, to show how good the current state is. 
The output actions of the policy network are then employed to update the segmentation results. 
Based on the changes of segmentation probability, the boundary-aware reward is proposed to measure the quality of the current action and further updates these parameters of the two networks. 
Meanwhile, the value map is used auxiliarly to update the parameters of the policy network.
The updated segmentation results are shown to experts for feedback. We design a new form of user interaction, \ie, supervoxel-clicking, which directs users to click on a point, suggesting that the corresponding supervoxel is wrongly predicted.
The clicks are later represented as interaction maps through geodesic transformation, and then used as part of the state in the next iteration.
Generally, the proposed BS-IRIS can iteratively refine the segmentation probability with experts' interactions until satisfying results are reached.

\subsection{MARL Framework of Interactive Image Segmentation}
\label{subsec:Multi-agent RL framework for interactive image segmentation}

Let $\bm{x} = (x_1, \cdots, x_N)$ be a 3D image, where $x_{i}$ is the $i^{th}$ voxel of $\bm{x}$ ($i=1,\cdots,N)$ with $N=H \times W \times C$. 
We treat each $x_{i}$ as an agent whose policy is denoted as $\pi_{i} (a_i^{(t)}|s_i^{(t)})$.  For $x_{i}$ at the iteration $t$, $s_i^{(t)} \in \mathcal{S}$ and $a_i^{(t)} \in \mathcal{A}$ denote the state and action, where $\mathcal{S}$ and $\mathcal{A}$ are the state space and the action space, respectively. 
The objective of BS-IRIS is to learn the optimal policy $\bm{\pi}=\left(\pi_{1}, \cdots, \pi_{N}\right)$ that maximizes the mean of the expected total discounted rewards for all voxels:
\begin{align}
\boldsymbol{\pi}^{*} &= \underset{\boldsymbol{\pi}}{\operatorname{argmax}} \  \mathbb{E}_{\boldsymbol{\pi}}\left(\sum_{t=0}^{T} \gamma^{t} \bar{r}^{(t)}\right),\\
\bar{r}^{(t)} &= \frac{1}{N} \sum_{i=1}^{N} r_{i}^{(t)},
\end{align}where $\gamma$ denotes the discount factor, $T$ denotes the total number of iterations, $N$ denotes the number of agents, and $\bar{r}^{(t)}$ denotes the mean of the rewards $r_{i}^{(t)}$ for all voxels.

The settings of BS-IRIS are slightly different from typical MARL~\cite{zhang2019multi, hernandez-leal2019a} in the following two aspects. 1) The number of agents $N$ is extremely large (in the scale of $10^5$ even after image resizing). It is impractical to directly apply typical multi-agent learning algorithms such as~\cite{lowe2017multi-agent}. 2) All voxel-type agents are neatly aligned in a 3D grid and dependent with each other for the segmentation task. It is necessary to require them to cooperate with each other. This is achieved by enforcing all voxel agents to share the same policy. When one agent explores a beneficial action, other agents will simultaneously acquire that knowledge, which also significantly reduces the number of parameters. 
The asynchronous advantage actor-critic (A3C)~\cite{mnih2016asynchronous} scheme is employed in BS-IRIS and further extended to a fully convolutional form, where agents can collaborate and communicate through convolutional layers. 
The algorithm is summarized in Alg.~\ref{alg}.

\begin{algorithm}
    \caption{Pseudo-code of BS-IRIS}
    \label{alg}
    \begin{algorithmic}
        \STATE // \textit{Assume global shared parameter vectors $\theta_p$, $\theta_v$, and global counter $T = 0$}
        \STATE // \textit{Assume thread-specific parameter vectors $\theta'_p$, $\theta'_v$}
        \STATE Initialize thread step counter $t = 1$
        \REPEAT
            \STATE Reset gradients: $d\theta_p = 0$, $d\theta_v = 0$
            \STATE Synchronize thread-specific parameters $\theta_p'=\theta_p$, $\theta_v'=\theta_v$
            \STATE $t_{start}=t$
            \REPEAT
                \STATE $T= T+1$, and $t= t+1$
                \STATE Interact with users, and get $h_{+,i}^{(t)}, h_{-,i}^{(t)}$ for $\forall i$
                \STATE Obtain $s_{i}^{(t)}=[v_{i}, p_{i}^{(t)}, h_{+,i}^{(t)}, h_{-,i}^{(t)}]$ for $\forall i$
                \STATE Perform $a_i^{(t)}$ according to  $\pi(a_i^{(t)}|s_i^{(t)};\theta_p')$ for $\forall i$
                \STATE $p_{i}^{(t+1)} = \mathcal{C}_0^1(p_{i}^{(t)} + a_{i}^{(t)})$, where $\mathcal{C}_{\alpha}^{\beta}(\cdot) = \min(\max(\cdot, \alpha), \beta)$  for $\forall i$
                \STATE Receive $r_i^{(t)}= r_{i,g}^{(t)} + \lambda  r_{i,b}^{(t)} $  for $\forall i$
            \UNTIL $t-t_{start}==t_{max}$
            \STATE $R_i = 0$  for $\forall i$
            \FOR{$k \in \{t-1,\cdots,t_{start}\}$}
            \STATE $R_i = {r}_i^{(k)}+ \gamma R_i$
            \STATE Accumulate gradients w.r.t. $\theta'_p$: $d\theta_p= d\theta_p-\nabla_{\theta_p'}{\frac{1}{N}\sum_{i=1}^N}\log\pi(a_i^{(k)}|s_i^{(k)};\theta_p')(R_i-V(s_i^{(k)};\theta_v'))$
            \STATE Accumulate gradients w.r.t. $\theta'_v$: $d\theta_v= d\theta_v+\nabla_{\theta_v'}{\frac{1}{N}\sum_{i=1}^N}(R_i-V(s_i^{(k)};\theta_v'))^2$
            \ENDFOR
            \STATE Update $\theta_p$, $\theta_v$ using $d\theta_p$ and $d\theta_v$  respectively
        \UNTIL $T>T_{max}$
    \end{algorithmic}
\end{algorithm}

\subsection{Supervoxel-level Interaction}

To combine the advantages of the point-clicking and the scribbles, we here propose a novel supervoxel-clicking interaction. 
While we name it  supervoxel-level interaction, the user interaction actually is the same as point-level interaction, \ie, clicking on the error positions. The algorithm simply automatically expand the point clicked to its corresponding supervoxel, so that it is not necessary to present the supervoxel to users.
Similar to point-clicking, the user only needs to click on one or multiple voxels to refine. 
As a supervoxel is a group of voxels sharing common characteristics, ``supervoxel-clicking'' has the advantages of providing more information in spatial context than point-clicking, as well as being more robust to random variations in clicking positions. 
This procedure iterates to improve the segmentation until satisfying performance is reached. 
We set the initial supervoxels size as $K_s$ and gradually shrink it by iteration.
The clicked supervoxels are denoted as the hint set $\mathbb{H}$.
Following~\cite{wang2018deepigeos}, user interactions can be categorized into object corrections and background corrections, so two form of hint sets are provided, denoted as $\mathbb{H}_{+}$ for objects and $\mathbb{H}_{-}$ for background respectively.

\subsection{State}
\label{subsec:state}
At each refinement iteration, the state is defined by the image itself, previous segmentation, and user interaction. Here the key is how to represent the interaction.
When giving the hint through a single click, the user indicates that the corresponding area is one of the error regions. 
To propagate the supervoxel-level interaction to the entire image, 
we here introduce \emph{interaction map} $\bm{h}$, which is of the same size as the image, to model the radiation of the user hints. 
Intuitively, the closer to the supervoxel that is being clicked, the more likely a voxel's label is incorrect. 
At each iteration, the $i^{th}$ element of the interaction map $h_{i}$ is thus calculated as the minimum distance from the corresponding voxel $x_i$ to voxels in the hint set $\mathbb{H}$:
$h_{i}=\min_{x_j \in \mathbb{H}} \mathcal{M}(x_i, x_j)$, where $\mathcal{M}$ is a function measuring the distance between two voxels. The geodesic distance is employed here to improve label consistency in homogeneous regions~\cite{wang2018deepigeos}. 

The geodesic distance of two voxels $x_i$ and $x_j$ is calculated by $ \mathcal{D}_{geo}(x_i, x_j) = \min _{p \in \mathcal{P}_{x_i, x_j}} \int_{0}^{1}\|\nabla \bm{x}(p(w)) \cdot \bm{u}(w)\| d w $, where $\mathcal{P}_{x_i, x_j}$ is the set of all paths between $x_i$ and $x_j$ ($x_i$, $x_j \in \mathbb{H}$) and $p$ is one feasible path parameterized by $w \in[0,1]$. $\bm{x}$ means the 3D image tensor with three coordinates. $\bm{u}(w)=p^{\prime}(w) /\left\|p^{\prime}(w)\right\|$ is a unit vector that is tangent to the direction of the path.
We compute the geodesic distance  with raster scan based algorithm~\cite{toivanen1996new} to leverage its kernel operations which can be sequentially applied over images~\cite{criminisi2008geos}.

Given the two hint sets $\mathbb{H}_{+}$ and $\mathbb{H}_{-}$, 
two corresponding interaction maps, $\bm{h}_{+}$ and $\bm{h}_{-}$ are generated.
For the voxel agent $x_{i}$, its state $s_{i}$ is the concatenation of its voxel value $v_{i}$, its previous segmentation probability $p_{i}$  and its two corresponding interaction map values $h_{+,i}$ and $ h_{-,i}$: $s_{i}=[v_{i}, p_{i}, h_{+,i}, h_{-,i}]$. The initial probability of $p_{i}$ is set to $0.5$.

\subsection{Action}
\label{subsec:action}
While some previous works directly output the segmentation probability from the network~\cite{bredell2018iterative,wang2018deepigeos}, to make the result more stable without abrupt changes, we here predict the adjustment amount based on the previous probability as actions.
The action $a \in \mathcal{A}$ is to adjust the segmentation probability $p$ by the amount $a$. The action set $\mathcal{A}$ contains $K$ actions, allowing the agent to learn adjusting the probability to various degrees under different situations. 
For the $i^{th}$ voxel, 
its segmentation probability $p_{i}^{(t+1)}$ after taking the action $a_{i}^{(t)}$ at  iteration $t$ is:
\begin{align}
p_{i}^{(t+1)} &= \mathcal{C}_0^1(p_{i}^{(t)} + a_{i}^{(t)}),
\end{align}
where $\mathcal{C}_{\alpha}^{\beta}(\cdot)=\min(\max(\cdot, \alpha), \beta)$ and $\mathcal{C}_{\alpha}^{\beta}(\cdot)$ clips the value from $\alpha$ to $\beta$. In this way, $p_{i}^{(t+1)}$ is constrained to $[0,1]$ since it represents a probability.

\subsection{Reward}
\label{subsec:reward}

To improve the efficiency of exploration and particularly considering the segmentation boundary which tends to be mis-segmented, we design a \textit{boundary-aware reward} consisting of the {\textit{global reward}} and the {\textit{boundary reward}}.

\medskip
\noindent {\textbf{Global reward}}. The global reward is the main indicator to evaluate the agent's actions. 
In order to explicitly consider the relation between successive predictions and to improve the efficiency of exploration, 
for the $i^{th}$ agent, its reward  $r_{i,g}^{(t)}$ at the $t^{th}$ iteration is designed as the relative improvement from the previous segmentation to the current one:
\begin{equation}
r_{i,g}^{(t)}  =   \chi_{i}^{(t-1)} - \chi_{i}^{(t)}, 
\end{equation}
where $\chi_{i} = -y_{i}\log(p_{i})-(1-y_{i})\log(1-p_{i})$ is the cross entropy loss and 
$y_{i}$ is the ground truth, and $p_{i}^{(t)}$ is the segmentation probability at c $t$.
With the global reward, the agent gets a positive reward when its prediction moves closer towards the true voxel label and vice versa. 

\medskip
\noindent {\textbf{Boundary reward}}. The global reward considers all agents equally in the ways of giving rewards and updating parameters. Considering that the boundary areas are often more error-prone, we introduce the boundary reward to make the agent more sensitive to the boundary during segmentation.
All agents are weighted according to their distance of the ground truth boundary, \ie, the agents near the boundary have higher weights than the others, so that they gain more rewards when doing things right, and are punished more otherwise.

Suppose $\mathcal{B}_{gt} = (b_1, b_2, \cdots, b_{M})$ is the ground truth boundary set of total $M$ voxels.
Without using too many symbols, we still use $x_i$ and $b_j$ to denote the specific voxels with 3D coordinate.
The weight for $x_i$ is defined as:
\begin{equation}
\mathcal{G}_i  =  \min_{b_j \in \mathcal{B}_{gt}} \|x_i- b_j \|_2.
\end{equation}
Denote the weight tensor as $\bm{\mathcal{G}} = (\mathcal{G}_1, \cdots, \mathcal{G}_N)$, $N = H \times W \times C$. The absolute boundary reward $\psi$ is formulated as:
\begin{equation}
\label{boudary_reward}
\psi_i^{(t)}=
\begin{cases}
+\mathcal{T}(\mathcal{G}_i) & \text{if $x_i$ is well predicted in iteration $t$}\\
-\mathcal{T}(\mathcal{G}_i) & \text{if $x_i$ is wrongly predicted in iteration $t$},
\end{cases}
\end{equation}
where $\mathcal{T}$ is an adaptive transformation. Here we choose:
\begin{equation}
\mathcal{T}(\mathcal{G}_i)  =  1- \frac{\mathcal{G}_i-\min(\bm{\mathcal{G})}}{\max(\bm{\mathcal{G}}) -\min(\bm{\mathcal{G}})}.
\end{equation}
Through a negative min-max normalization to $[0,1]$, the monotonicity is changed to ensure the largest value is close the boundary, and lowers the weight variance between images.
Intuitively, with the formulation of  $\psi_i^{(t)}$, the agent gets a positive reward when predicted correctly, and a negative reward (punishment) otherwise, and the  reward or punishment is higher when it is closer to the ground truth boundary.

The final boundary reward $r_{i,b}$ is the relative improvement of previous $\psi_i^{(t-1)}$ and current $\psi_i^{(t)}$ like the global reward:
\begin{equation}
r_{i,b}^{(t)}  =  \psi_i^{(t)} - \psi_i^{(t-1)}.  
\end{equation}

\medskip
\noindent {\textbf{Overall reward}}. Combining the global reward and the boundary reward, we have 
$r_i^{(t)}  =  r_{i,g}^{(t)} + \lambda  r_{i,b}^{(t)}$, 
where $\lambda$ controls the weight of the boundary reward.

In general, the accumulated reward for agent $x_i$ of one interactive sequence is:
\begin{equation}
\label{sequence reward}
\begin{aligned}
R_i  
& 
 =  \sum_{t=1}^{T} \gamma^{t-1} \left(r_{i,g}^{(t)} + \lambda r_{i,b}^{(t)}\right) \\
&  =  R_{i,g} + \lambda  R_{i,b},
\end{aligned}
\end{equation}
where $T$ is the total iteration number, and the discount factor $\gamma$ takes a value in $(0, 1]$.

\subsection{Network and Training}

\begin{figure}
\centering
\includegraphics[width=3.6in]{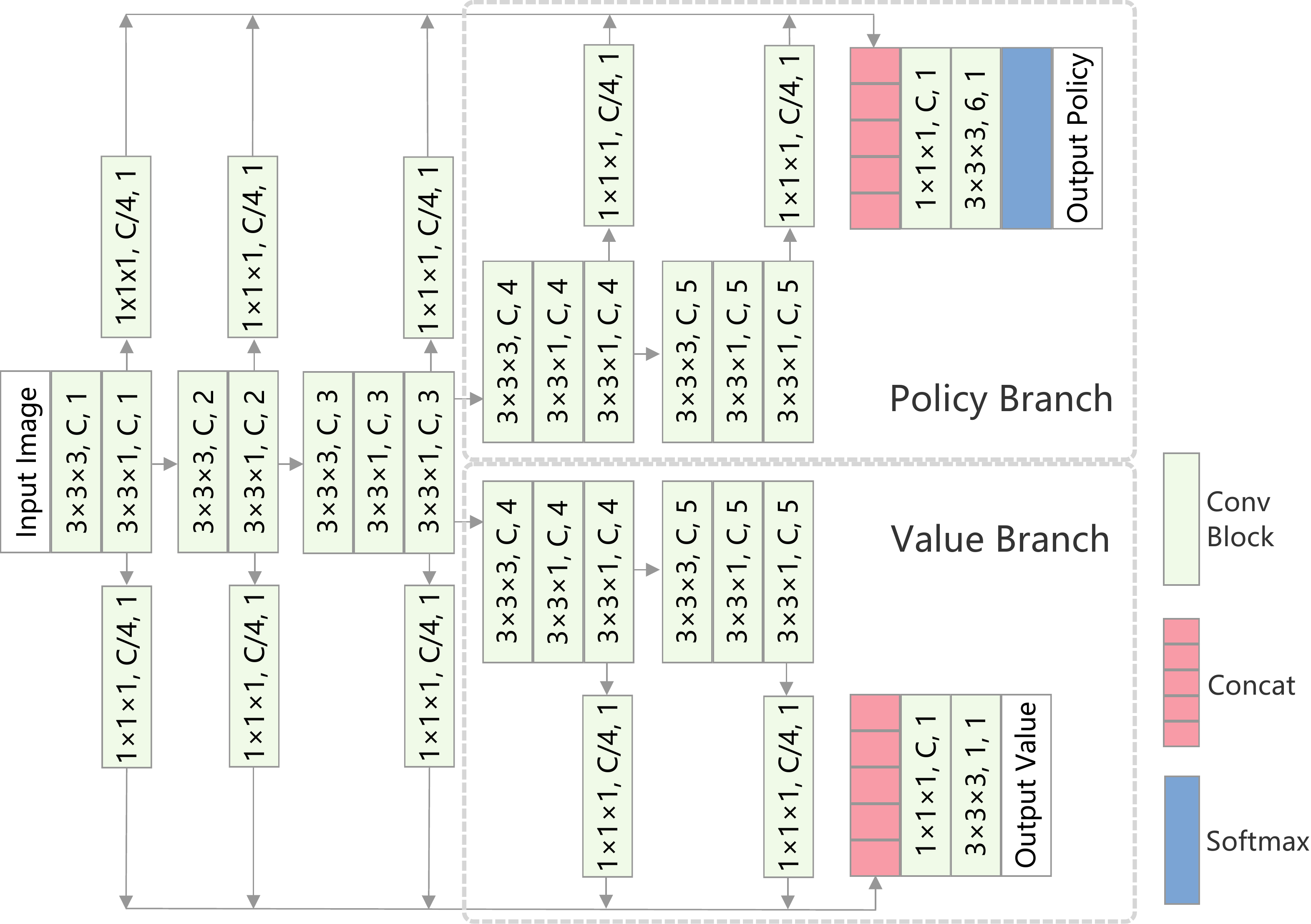}
\caption{The network architecture for BS-IRIS. The policy and value branches share the low-level features and extract their own high-level features. The green rectangle is a convolutional layer followed by ReLU, the text on which represents the kernel size, the number of output channel, and the dilation rate of atrous convolution separately. The red rectangle is the concatenation of multi-scale features. The blue rectangle is the softmax operation.}
\label{architecture}
\end{figure}

We employ asynchronous advantage actor-critic (A3C)~\cite{mnih2016asynchronous}  and extend it to a fully convolutional form. The network architecture of~\cite{wang2018deepigeos} is adopted and modified to fit as the backbone, as shown in Fig. \ref{architecture}. 
The network firstly uses three 3D convolutional blocks to extract low-level features. 
Then, the network is divided into two parts: policy branch and value branch, both of which have three 3D convolutional blocks to extract specific high-level features. 
The functionality of the policy branch is to predict the distribution of action probabilities under a known state. In our case, given the image, interaction maps, and previous segmentation probability, the policy branch predicts how likely it is to take each scale of adjustment to the previous segmentation probability. 
The functionality of the value branch is to estimate the value of the current state. Specifically, the value branch evaluates how good the current combination of the image, interaction maps, and the previous segmentation probability is.

We respectively use $\theta_p$ and $\theta_v$ to denote the parameters of the policy and value branches. The input of the network is the state ${s}^{(t)}$ at iteration $t$. The value branch outputs the estimated value of the current state ${V}({s_i}^{(t)})$ for all agents. The gradient for $\theta_v$ is computed by:
\begin{equation}
d\theta_v = \nabla_{\theta_v}{\frac{1}{N}\sum_{i=1}^N} A({s_i}^{(t)}, {a_i}^{(t)})^2, 
\end{equation}
\begin{equation}
A({s_i}^{(t)}, {a_i}^{(t)})  = \sum_{k=t}^{T}\gamma^{k-t} {r_i}^{(k)} - V(s_i^{(t)}),
\end{equation}
where ${r_i}^{(k)}$ is the reward for voxel agent $i$ at iteration $k$. $A({s_i}^{(t)}, {a_i}^{(t)})$ is the advantage of agent $i$ at iteration $t$ when taking ${a_i}^{(t)}$ in condition of state ${s_i}^{(t)}$, which indicates the accumulated reward without being affected by the state and reduces the variance of gradient. The policy branch outputs the policy $\pi({a_i}^{(t)}|{s_i}^{(t)})$ for all agents as the probabilities of taking each action ${a_i}^{(t)}$. The gradient for $\theta_p$ is computed by:
\begin{equation}
d\theta_p = -\nabla_{\theta_p}{\frac{1}{N}\sum_{i=1}^N}\log\pi(a_i^{(k)}|s_i^{(k)}) A({s_i}^{(t)}, {a_i}^{(t)}).
\end{equation}
The two branches are jointly trained in an end-to-end manner.

\section{Experiments}

\subsection{Datasets and Evaluation Metrics}
We evaluate our algorithm on four publicly available benchmark medical image segmentation datasets. 
\begin{itemize}[leftmargin=12pt,itemsep=0pt,parsep=0pt]
\item \textbf{BraTS2015~\cite{menze:hal-00935640}(training set).} We use FLAIR images and segment the whole tumor (WT), tumor core (TC) and enhancing tumor (ET) separately. $234$ cases are randomly selected for training and the remaining $40$ for test.
\item \textbf{BraTS2019~\cite{bakas2018identifying}(training set).}  We use FLAIR images and segment the whole tumor (WT), tumor core (TC) and enhancing tumor (ET) separately. $285$ cases are randomly selected for training and the remaining $50$ for test.
\item \textbf{MM-WHS~\cite{zhuang2016multi}(training set).}
We only use MRI and only segment left atrium blood cavity (LA). 5-fold cross validation is conducted on the total $20$ cases.
\item \textbf{NCI-ISBI2013~\cite{bloch2015nci}.} We segment peripheral zone (PZ) and central gland (CG). $40$ cases are randomly selected for training and the remaining $20$ for test.
\end{itemize}
To evaluate the interactive segmentation methods, we employ three metrics under the same interaction settings: Dice similarity coefficient (DSC), average symmetric surface distance (ASSD) and $95^{th}$ percentile Hausdorff distance (HD95). Students's $t$-test is used to compute the $p$-value to show whether the results of two algorithms are significantly different.

\subsection{User Interactions Simulation}
\label{subsec:User interaction and simulation}

For the supervoxel-level interaction, it is demanding to interact with real physicians. A robot user is thus utilized to simulate user clicks following related studies~\cite{wang2018deepigeos, bredell2018iterative}. The clicking position of a simulated click is set at the center of the biggest mis-segmented supervoxel (\ie, the supervoxel with the largest error region in all supervoxels).
In one interaction sequence of each case, each iteration receives $N_{c}$ clicks.
Besides, a small disturbance $\epsilon$ is added to each clicking position to imitate the behavior of a real user.

\subsection{Implementation Details}

As pre-processing, each image is normalized by its mean and standard variation, and is cropped based on the region of interest. Data augmentation involves flipping and random rotation with angle range $[-\pi/8, \pi/8]$ both in three directions.

As BS-IRIS can easily adapt to the interaction-free mode, we first train a warm-up model by training without any interactions for about $100$ to $200$ epochs. 
We then employ the simulated clicks by using the warm-up model as startup for about $500$ epochs with $4$ interactions ($6$ clicks/interaction). 
Parameter setting is as follows: total iterations $T=4$, clicks per iteration $N_{c}=6$, $\gamma=0.95$, $\epsilon=[-3,3]^3$, network channel $C=64$. 
We use Adam algorithm for optimization with minibatch size $1$.
The learning rate adopts the iteration decay schedule with initialized to $10^{-4}$. 

For the supervoxel segmentation, we use SLIC~\cite{achanta2010slic} with $spacing=[2,2,2]$, $compactness=0.1$, and the number of initial supervoxels equals to $100$ and gradually declines during the refinement iterations for training and testing.

The training time with one Nvidia Tesla V100 GPU varies from several hours to a few days on different datasets. The average inference time for each update iteration is about $990$ms, including about $280$ms for supervoxel segmentation and $250$ms for geodesic transformation calculation. It thus takes about $4$s to annotate a 3D image with $4$ interactive iterations. For real-world application, model compression techniques may be further leverage to speed up the process. But it is out of the scope of this paper. 
\section{Results and Analysis}

\subsection{Ablation Study}

We perform a comprehensive ablation study to evaluate each component in our multi-agent reinforcement learning method for interactive segmentation. Without any explicit mention, all the ablation experiments are performed with the whole tumor (WT) segmentation task of BraTS2015.

\subsubsection{Effect of state and action settings}

\begin{table}
    \centering
    \caption{The Influence of Action and State Settings on BraTS2015 with 4 Interactions (6 Clicks/Interaction). Significant Improvement ($p\text{-value} < 0.05$) Is Marked in Bold}
        \begin{threeparttable}
    	\begin{tabular}{c c c c | c c | c}
            \toprule[1pt]
            \multicolumn{4}{c|}{Action} & \multicolumn{2}{c|}{State} & \multirow{2}*{DSC(\%)$\uparrow$}                                                          \\ \cline{1-6}
            $\pm$0.1                     & $\pm$0.2                    & $\pm$0.4               & $\pm$1.0   & Probability & Binary     &                \\ 
            \hline
                                         &                             &                        & \checkmark &             & \checkmark & - \tnote{*}          \\
                                         &                             & \checkmark             &            & \checkmark  &            & 86.35$_{\pm4.14}$          \\
                                         & \checkmark                  &                        &            & \checkmark  &            & 88.69$_{\pm3.92}$          \\
            \checkmark                   &                             &                        &            & \checkmark  &            & 89.34$_{\pm3.75}$          \\
            \checkmark                   & \checkmark                  &                        &            & \checkmark  &            & 90.02$_{\pm3.96}$          \\
            \checkmark                   & \checkmark                  & \checkmark             &            & \checkmark  &            & \textbf{90.81}$_{\pm3.42}$ \\
            \checkmark                   & \checkmark                  & \checkmark             & \checkmark & \checkmark  &            & 90.25$_{\pm3.72}$          \\
            \bottomrule[1pt]
        \end{tabular}
    \label{tab:state and action}
        \begin{tablenotes}
    	\footnotesize
    	\item[*]  Hard to converge.
    \end{tablenotes}
\end{threeparttable}
\end{table}

We only consider discrete actions in our study because continuous action space makes the model difficult to train and converge. When the action set only contains ${\pm1.0}$, the segmentation probability becomes binary. The loss of prediction uncertainty makes the model unable to converge. The rest of the action sets are thus all designed for the states enabling segmentation probability.

Table \ref{tab:state and action} shows the performance of different action sets. 
The first four rows fix the number of actions to be one and compare different action values. 
It is observed that small action values lead to better performances than the larger ones. The reason is that a small action value allows the model to make more fine-grained adjustments, but a larger one may over-act and can not guarantee to converge to the optimal results. 

For the influence of the number of actions, we gradually add new actions to the action set. It can be observed  that abundant actions lead to better performance by providing the model with various degrees of adjustment. In the case with high certainty, the model tends to take large adjustments, thereby accelerating the refinement convergence. However, the addition of $\pm1.0$ relatively damages the performance for an adjustment of $\pm1.0$ is too extreme in most situations.

Based on the above experiments, we finally choose the final action set as $\mathcal{A}=\{\pm0.1, \pm0.2, \pm0.4\}$.  

\begin{table}
    \centering
    \caption{Comparison of Different Reward Settings on BraTS2015 with 4 Interactions (6 Clicks/Interaction). Significant Improvement ($p\text{-value} < 0.05$) Is Marked in Bold}
    \setlength{\tabcolsep}{3pt}
    \begin{threeparttable}
        \begin{tabular}{cc|ccc}
            \toprule[1pt]
            \multicolumn{2}{c|}{Reward} & DSC(\%)$\uparrow$       & ASSD(pixels)$\downarrow$               & HD95(mm)$\downarrow$                                              \\  
            \hline
            \multicolumn{1}{c|}{\multirow{2}{*}{Absolute}}   & Global         & 88.67$_{\pm3.81}$          & 1.49$_{\pm1.01}$          & 4.21$_{\pm3.16}$          \\ 
            \multicolumn{1}{c|}{}          & Boundary-aware & 89.59$_{\pm3.56}$          & 1.30$_{\pm1.19}$          & 3.90$_{\pm3.01}$          \\ 
            \hline
            \multicolumn{1}{c|}{\multirow{2}{*}{Relative}}   & Global         & 89.80$_{\pm3.57}$          & 1.25$_{\pm0.88}$          & 3.87$_{\pm3.21}$          \\ 
            \multicolumn{1}{c|}{}         & Boundary-aware & \textbf{90.81}$_{\pm3.42}$ & \textbf{1.05}$_{\pm0.53}$ & \textbf{3.07}$_{\pm1.39}$ \\ 
            \bottomrule[1pt]
        \end{tabular}
        \label{tab:reward}
    \end{threeparttable}
\end{table}

\subsubsection{Effect of different rewards}

For each reward, in addition to the relative formation, we also consider its absolute counterpart, \ie, $r_i =-\chi_{i} + \lambda \psi_i$.
As can be seen from Table \ref{tab:reward}, for both the 
absolute and relative formulation of the rewards, the boundary-aware rewards result in better performance than the global rewards only, suggesting that weighting more on the boundaries helps improve the segmentation performance. 
Compared to the absolute rewards, the relative rewards lead to better performance. Because joining a previous performance as a baseline, the model has a target to compare. The relative reward is shown to improve space exploration efficiency in RL policy, which adds a limit and direction for model updating.

\subsubsection{Effect of different forms of interactions}

\begin{table}
    \centering
    \caption{Quantitative Comparison of Different Interaction Types (with Different Reward Types) on BraTS2015 with 4 Interactions (6 Clicks/Interaction). Significant Improvement ($p\text{-value} < 0.05$) Is Shown in Bold}
    \setlength{\tabcolsep}{3pt}
        \begin{tabular}{c|ccc}
            \toprule[1pt]
            {Interaction (reward type)}        & DSC(\%)$\uparrow$    & ASSD(pixels)$\downarrow$               & HD95(mm)$\downarrow$            \\  
            \hline
            Point (global)     & 89.45$_{\pm3.66}$          & 1.27$_{\pm0.99}$          & 3.98$_{\pm3.27}$          \\ 
            Supervoxel (global)  & 89.90$_{\pm3.57}$          & 1.25$_{\pm0.88}$          & 3.87$_{\pm3.21}$          \\ 
            \hline
            Point (boundary-aware)     & 89.62$_{\pm3.61}$          & 1.24$_{\pm0.74}$          & 3.89$_{\pm2.48}$          \\ 
                     Supervoxel (boundary-aware)  & \textbf{90.81}$_{\pm3.42}$ & \textbf{1.05}$_{\pm0.53}$ & \textbf{3.07}$_{\pm1.39}$ \\
            \bottomrule[1pt]
        \end{tabular}
\label{tab:interaction types}
\end{table}

We experiment with different interaction types, \ie, point-clicking, and supervoxel-clicking. As can be seen from Table \ref{tab:interaction types}, with the same numbers of clicking, supervoxel-clicking performs better than point-clicking, for both the global reward and the boundary-aware reward. With the same clicking numbers, supervoxel represents a clustering of points and provides more information than only one point.

\subsubsection{Effect of interaction quality}
\begin{table}
    \centering
    \caption{Comparison of Different Interaction Quality on BraTS2015 with 4 Interactions (6 Clicks/Interaction). Significant Improvement ($p\text{-value} < 0.05$) Is Marked in Bold for Supervoxel-clicking, and Is Underlined for Point-clicking}
    \setlength{\tabcolsep}{3pt}
    \begin{tabular}{cc|c|ccc}
        \toprule[1pt]
    \multicolumn{2}{c|}{Quality}                                           & Type & DSC(\%)$\uparrow$ & ASSD(pixels)$\downarrow$ & HD95(mm)$\downarrow$ \\ \hline
    \multicolumn{2}{c|}{No Interaction}                                                 & n/a           & 85.47$_{\pm7.10}$             & 1.87$_{\pm1.32}$             & 7.17$_{\pm4.71}$      \\ \hline
    \multicolumn{1}{c|}{\multirow{8}{*}{\tabincell{c}{Noise \\ Range}}} &

    \multirow{2}{*}{0}   & point        & 89.45$_{\pm3.66}$ & \underline{1.27}$_{\pm0.99}$ & \underline{3.98}$_{\pm3.27}$      \\ 
    \multicolumn{1}{c|}{}                       &                               & supervoxel   & \textbf{90.81}$_{\pm3.42}$    & 1.05$_{\pm0.53}$             & \textbf{3.07}$_{\pm1.39}$    \\ \cline{2-6} 
    \multicolumn{1}{c|}{}                       &
    \multirow{2}{*}{{[}-3,3{]}}   & point        & \underline{89.46}$_{\pm3.75}$ & 1.29$_{\pm1.02}$             & 4.00$_{\pm3.72}$      \\ 
    \multicolumn{1}{c|}{}                       &                               & supervoxel   & {90.79}$_{\pm3.46}$           & \textbf{1.02}$_{\pm0.85}$    & 3.08$_{\pm1.40}$      \\ \cline{2-6} 
    \multicolumn{1}{c|}{}                       & \multirow{2}{*}{{[}-5,5{]}}   & point       & 88.09$_{\pm4.12}$             & 1.53$_{\pm1.09}$             & 5.04$_{\pm4.02}$               \\ 
    \multicolumn{1}{c|}{}                       &                               & supervoxel  & 90.60$_{\pm3.99}$             & 1.27$_{\pm0.99}$             & 3.16$_{\pm3.61}$               \\ \cline{2-6} 
    \multicolumn{1}{c|}{}                       & \multirow{2}{*}{{[}-7,7{]}} & point       & 86.53$_{\pm4.50}$             & 1.82$_{\pm1.16}$             & 6.64$_{\pm4.59}$               \\
    \multicolumn{1}{c|}{}                       &                               & supervoxel  & 89.68$_{\pm4.39}$             & 1.44$_{\pm1.04}$             & 3.74$_{\pm4.53}$               \\ \hline
    \multicolumn{2}{c|}{\multirow{2}{*}{Random}}                                 & point       & 84.52$_{\pm7.11}$             & 1.96$_{\pm1.42}$             & 7.81$_{\pm4.97}$           \\  
    \multicolumn{2}{c|}{}                                                        & supervoxel  & 84.23$_{\pm6.98}$             & 2.01$_{\pm1.58}$             & 8.02$_{\pm5.20}$           \\ \bottomrule[1pt]
    \end{tabular}
    \label{tab:quality}
    \end{table}

The clicking position from real-world users may not be as ideal as the simulation. We further validate the robustness of BS-IRIS under noisy click positions and report the experimental results in Table \ref{tab:quality}. 
Here ``No interaction'' means the results obtained by inputting empty clicks (corresponding to all zero tensor for the two interaction maps), ``Noise Range'' represents degree of perturbations in simulating the clicks, and ``Random'' means to randomly choose clicks.
For the three metrics, both two types of interaction are tolerant to noise when the noise range is small, but their performance starts to decline as noise range increases. The supervoxel-clicking is more robust as it leads to more stable results across different noise ranges, even the range is larger than $[-3,3]$.
The reason is that adding a bias on clicking position usually does not  change the choice of supervoxels as the supervoxel covers a reasonably large region. As long as the click position is still in the range of the supervoxels, the result will not be affected.
We also experiment with randomly selected clicks. 
Compared to the models trained with no interactions, random clicks lead to even worse results. 

\subsubsection{Effect of interaction map generations}

\begin{table}
    \centering
    \caption{Comparison of Interaction Map Generation Functions on BraTS2015 with 4 Interactions (6 Clicks/Interaction). Significant Improvement ($p\text{-value} < 0.05$) Is Marked in Bold}
    \begin{threeparttable}
        \begin{tabular}{c| c c c}
                \toprule[1pt]
                Generation Function  & DSC(\%)$\uparrow$                    & ASSD(pixels)$\downarrow$      & HD95(mm)$\downarrow$  \\ 
                \hline
                Euclidean & 90.29$_{\pm3.58}$          & 1.09$_{\pm0.68}$          & 3.72$_{\pm2.14}$          \\ 
                Gaussian  & 89.57$_{\pm3.27}$          & 1.13$_{\pm0.51}$          & 4.03$_{\pm2.19}$          \\ 
                Geodesic  & \textbf{90.81}$_{\pm3.42}$ & \textbf{1.05}$_{\pm0.53}$ & \textbf{3.07}$_{\pm1.39}$ \\ 
                \bottomrule[1pt]
            \end{tabular}
        \label{tab:distance}
    \end{threeparttable}
\end{table}

We evaluate three distance metrics, \ie,  the Euclidean, Gaussian, and geodesic distances, for interaction maps generation.
Table \ref{tab:distance} presents the results of the interaction maps based on the three distance metrics.
Among the three, the geodesic distance leads to the best performance, the Euclidean distance comes in second, and Gaussian distance performs worst. The computation of geodesic distance not only involves the absolute distance between voxels, but also integrates the low-level image features of the corresponding path. Its superior results among the threes suggest that the information of the image itself is also important for generating the interaction map. On the other hand, the Gaussian distance performs worse than Euclidean distance, possibly because its influence of clicking point declines rapidly within a limited range.
Although such a conservative approach tends to avoid wrong hints, it may have little influence on the final prediction through back propagation, since the original image dominates the segmentation results.

\begin{table}
	\centering
	\caption{Comparison of Different Interaction Patterns on BraTS2015. Significant Improvement ($p\text{-value} < 0.05$) Is Marked in Bold}
	\setlength{\tabcolsep}{3pt}
	\begin{threeparttable}
		\scalebox{1}{
			\begin{tabular}{c c| c c c}
				\toprule[1pt]
				 Click/Iteration & Iteration & DSC(\%)$\uparrow$ & ASSD(pixels)$\downarrow$ & HD95(mm)$\downarrow$ \\
				\hline
				 1 & 24 & 89.45$_{\pm3.01}$    & 1.31$_{\pm1.09}$    & 4.01$_{\pm3.17}$ \\ 
				 2 & 12  & 89.83$_{\pm2.98}$   & 1.24$_{\pm0.84}$     & 3.97$_{\pm2.25}$ \\ 
				 3 & 8  & 90.21$_{\pm3.29}$    & 1.20$_{\pm0.92}$    & 3.83$_{\pm2.71}$ \\ 
				 4 & 6  & \textbf{90.82}$_{\pm3.40}$    & 1.09$_{\pm0.49}$    & 3.10$_{\pm1.30}$ \\ 
				 6 & 4  & {90.81}$_{\pm3.42}$    & \textbf{1.05}$_{\pm0.53}$    & \textbf{3.07}$_{\pm1.39}$ \\ 
 				 8 & 3  & 90.35$_{\pm3.50}$    & 1.16$_{\pm0.79}$    & 3.81$_{\pm1.96}$ \\ 
				 12 & 2  & 89.61$_{\pm4.19}$    & 1.22$_{\pm1.19}$    & 3.86$_{\pm3.27}$ \\ 
				 24 & 1  & 89.33$_{\pm4.37}$    & 1.39$_{\pm1.32}$    & 4.06$_{\pm3.69}$ \\ 
				\bottomrule[1pt]
			\end{tabular}
		}
		\label{tab:step and click}
	\end{threeparttable}
\end{table}

\subsubsection{Effect of interaction patterns}

Given the total budget of interaction (\ie, $24$ clicks per image in our study), we experiment to find the optimal clicking numbers per iteration. Table \ref{tab:step and click} shows that the model reaches the optimum with $6$ clicks/iteration. A possible explanation is that, if the number of clicks is larger than the number of error regions in one iteration, most of the clicks will not be fully utilized. Conversely, small number of clicks could miss some error regions and  increase the total iteration.
Balancing the time and performance, we choose 6 clicks/iteration in our experiments.

\subsubsection{Effect of supervoxels region sizes}
The size of supervoxels region is important for our method. We experiment with four supervoxel sizes, including three fixed sizes and a declined schedule one. As shown in Table \ref{tab:supevoxel size}, the declined schedule has the best performance, for its trending is identical to the segmentation probability updating tendency, that is, as the iteration increased, the segmentation should be more fine-grained and supervoxels of smaller size should be adopted.

\begin{table}
	\centering
	\caption{Comparison of Different Supervoxel Region Sizes on BraTS2015 with 4 Interactions (6 Clicks/Interaction). Significant Improvement ($p\text{-value} < 0.05$) Is Marked in Bold}
	\setlength{\tabcolsep}{5pt}
	\begin{threeparttable}
			\begin{tabular}{c| c c c}
				\toprule[1pt]
				Supervoxel Region & DSC(\%)$\uparrow$             &ASSD(pixels)$\downarrow$ & HD95(mm)$\downarrow$ \\
				\hline
				Large ($size=100$)& 89.42$_{\pm3.89}$     & 1.28$_{\pm0.78}$     & 4.01$_{\pm2.02}$ \\ 
				Decline\tnote{*}  & \textbf{90.81}$_{\pm3.42}$    & \textbf{1.05}$_{\pm0.53}$    & \textbf{3.07}$_{\pm1.39}$ \\ 
				Small ($size=10$) & 90.57$_{\pm3.73}$    & 1.13$_{\pm0.66}$    & 3.13$_{\pm1.46}$ \\ 
				Point ($size=1$) & 89.62$_{\pm3.61}$    & 1.24$_{\pm0.74}$    & 3.89$_{\pm2.48}$ \\ 
				\bottomrule[1pt]
			\end{tabular}
		\label{tab:supevoxel size}
		
		\begin{tablenotes}
			\footnotesize
			\item[*] We choose $size=\lceil 100 \times 0.45^{{iteration}} \rceil$ in this declined situation. For the first iteration ($iteration=0$), the initial supervoxel region size is $100$, and the last $4^{th}$ iteration ($iteration=3$), the region size equals to $10$, which is corresponding to large and small sizes in this table respectively.   
		\end{tablenotes}
	\end{threeparttable}
\end{table}

\subsection{Comparisons with State-of-the-art Methods}
We compare BS-IRIS with four state-of-the-art interactive segmentation methods: Min-Cut~\cite{krahenbuhl2011efficient}, InterCNN~\cite{bredell2018iterative}, DeepIGeoS~\cite{wang2018deepigeos}, and IteR-MRL~\cite{liao2020iteratively}.

\subsubsection{Performances on different datasets}

\begin{table*}
    \centering
    \caption{Comparison of Interactive Segmentation Methods on BraTS2015 with 4 Interactions (6 Clicks/Interaction). Significant Improvement ($p\text{-value} < 0.05$) Is Marked in Bold}
    \begin{threeparttable}
        \begin{tabular}{c|ccc|ccc}
        \toprule[1pt]
        \multirow{2}{*}{Method} & \multicolumn{3}{c|}{DSC(\%)$\uparrow$} & \multicolumn{3}{c}{HD95(mm)$\downarrow$} \\ \cline{2-7} 
                                & Whole Tumor & Tumor Core & Enhancing Tumor & Whole Tumor & Tumor Core & Enhancing Tumor      \\ \hline
        State-of-the-art~\cite{2019One}\tnote{*} & 87    & 75   & 65   & -    & -   & -   \\ \hline
        Min-Cut                  & 82.67$_{\pm7.83}$     & 73.15$_{\pm7.79}$   & 63.43$_{\pm8.16}$  & 4.99$_{\pm4.39}$    & 6.91$_{\pm5.10}$   & 4.43$_{\pm4.00}$   \\
        InterCNN & 86.75$_{\pm7.23}$  & 75.10$_{\pm6.94}$   & 65.68$_{\pm6.81}$   & 4.24$_{\pm3.57}$    & 5.90$_{\pm3.80}$   & 3.81$_{\pm3.28}$   \\
        DeepIGeoS & 87.19$_{\pm6.34}$   & 76.26$_{\pm5.68}$   & 65.42$_{\pm7.42}$   & 4.12$_{\pm3.59}$    & 5.24$_{\pm3.63}$   & 3.97$_{\pm3.43}$   \\
        IteR-MRL & 89.45$_{\pm3.66}$  & 77.01$_{\pm3.42}$  & 66.01$_{\pm3.70}$   & 3.74$_{\pm3.16}$    & \textbf{5.15}$_{\pm3.10}$   & 3.39$_{\pm3.14}$   \\
        BS-IRIS & \textbf{90.81}$_{\pm3.42}$ & \textbf{77.98}$_{\pm3.58}$ & \textbf{66.69}$_{\pm3.39}$   & \textbf{3.22}$_{\pm3.02}$    & 5.19$_{\pm3.21}$   & \textbf{3.13}$_{\pm2.97}$   \\ 
        \bottomrule[1pt]
        \end{tabular}
        
        \label{tab:brats15}
        
        \begin{tablenotes}
            \footnotesize
            \item[*] 
            The model is trained on BraTS2015 training set and evaluated on the testing set (no HD95 results reported). Because we can no longer access the testing data, the results for the other methods are obtained on the training set only with a train/test split.
        \end{tablenotes}
        
    \end{threeparttable}
\end{table*}

\begin{table*}
	\centering
	\caption{Comparison of Interactive Segmentation Methods on BraTS2019 with 4 Interactions (6 Clicks/Interaction). Significant Improvement ($p\text{-value} < 0.05$) Is Marked in Bold}
	\begin{threeparttable}
		\begin{tabular}{c|ccc|ccc}
			\toprule[1pt]
			\multirow{2}{*}{Method} & \multicolumn{3}{c|}{DSC(\%)$\uparrow$} & \multicolumn{3}{c}{HD95(mm)$\downarrow$} \\ \cline{2-7} 
			& Whole Tumor & Tumor Core & Enhancing Tumor & Whole Tumor & Tumor Core & Enhancing Tumor \\ \hline
			State-of-the-art~\cite{jiang2019two}\tnote{*} & 88.80    & 83.70   & 83.27   & 4.62    & 4.13   & 2.65   \\ \hline
			Min-Cut & 87.96$_{\pm7.01}$   & 83.01$_{\pm7.19}$  & 82.09$_{\pm7.12}$  & 5.42$_{\pm4.93}$  & 4.40$_{\pm4.36}$   & 3.01$_{\pm3.24}$   \\
			InterCNN & 89.54$_{\pm5.90}$  & 85.17$_{\pm5.99}$  & 83.59$_{\pm5.97}$   & 4.17$_{\pm3.52}$ & 3.34$_{\pm3.25}$   & 2.74$_{\pm2.70}$   \\
			DeepIGeoS & 90.91$_{\pm5.58}$ & 85.64$_{\pm5.91}$  & 83.87$_{\pm5.89}$  & 3.40$_{\pm3.10}$  & 3.23$_{\pm2.98}$   & 2.58$_{\pm2.54}$   \\
			IteR-MRL & 91.27$_{\pm3.59}$  & 86.05$_{\pm3.62}$  & 84.15$_{\pm3.60}$  & 2.99$_{\pm2.35}$  & 3.10$_{\pm2.56}$   & 2.41$_{\pm2.39}$   \\
			BS-IRIS & \textbf{92.01}$_{\pm3.34}$ & \textbf{86.83}$_{\pm3.69}$  & \textbf{84.71}$_{\pm3.51}$ & \textbf{2.59}$_{\pm2.42}$    & \textbf{2.92}$_{\pm2.82}$   & \textbf{2.34}$_{\pm2.32}$   \\ \hline
			BS-IRIS$_{BraTS2015}$\tnote{**}  & 91.15 & 86.12 & 84.06  & 2.64  & 3.11  & 2.49 \\ 
			\bottomrule[1pt]
		\end{tabular}
		
		\label{tab:brats19}
		
		\begin{tablenotes}
			\footnotesize
			\item[*] 
			The model is trained on BraTS2019 training set and evaluated on the testing set. Because we can no longer access the testing data, the results for the other methods are obtained on the training set only with a train/test split.
			\item[**] This model is trained on BraTS2015.
		\end{tablenotes}
		
	\end{threeparttable}
\end{table*}

\begin{table}
	\centering
	\caption{Comparison of Interactive Segmentation Methods on MM-WHS with 4 Interactions (6 Clicks/Interaction). Significant Improvement ($p\text{-value} < 0.05$) Is Marked in Bold}
	\begin{threeparttable}
		\begin{tabular}{c |c c c}
			\toprule[1pt]
			Method & DSC(\%)$\uparrow$ & ASSD(pixels)$\downarrow$ & HD95(mm)$\downarrow$ \\
			\hline 
			State-of-the-art~\cite{2018Bayesian}\tnote{*} & 85.60       & -  &- \\
			\hline
			Min-Cut & 83.98$_{\pm4.78}$        & 1.70$_{\pm1.76}$          & 5.27$_{\pm4.83}$ \\ 
			InterCNN & 85.65$_{\pm1.98}$        & 1.42$_{\pm1.22}$          & 4.81$_{\pm3.51}$ \\ 
			DeepIGeoS & 86.01$_{\pm2.10}$       & 1.39$_{\pm1.25}$          & 4.31$_{\pm3.21}$ \\ 
			IteR-MRL & 87.19$_{\pm1.32}$        & 1.33$_{\pm0.59}$          & 4.10$_{\pm1.99}$ \\ 
			BS-IRIS & \textbf{89.00}$_{\pm1.12}$ & \textbf{1.24}$_{\pm0.53}$ & \textbf{3.71}$_{\pm1.59}$ \\ 
			\bottomrule[1pt]
		\end{tabular}
		\label{tab:mmwhs}
		
		\begin{tablenotes}
			\footnotesize
			\item[*] The model is trained on MM-WHS training set and evaluated on the testing set (no ASSD and HD95 results reported). Because we can no longer access the testing data, the results for the other methods are obtained on the training set only with a train/test split.
		\end{tablenotes}
	\end{threeparttable}
\end{table}

\begin{table}
    \centering
    \caption{Comparison of Interactive Segmentation Methods on NCI-ISBI2013 with 4 Interactions (6 Clicks/Interaction). Significant Improvement ($p\text{-value} < 0.05$) Is Marked in Bold}
    \begin{threeparttable}
		\begin{tabular}{c |c c c}
			\toprule[1pt]
            Method & DSC(\%)$\uparrow$                   & ASSD(pixels)$\downarrow$ & HD95(mm)$\downarrow$ \\
			\hline 
			State-of-the-art~\cite{20203D}\tnote{*} & 85.40       & -  &- \\
			\hline
			Min-Cut & 83.27$_{\pm4.90}$        & 1.55$_{\pm2.02}$          & 5.61$_{\pm5.06}$ \\ 
			InterCNN & 85.05$_{\pm3.47}$        & 1.19$_{\pm1.97}$          & 4.27$_{\pm4.19}$ \\ 
			DeepIGeoS & 84.95$_{\pm2.24}$       & 1.34$_{\pm1.01}$          & 4.70$_{\pm3.31}$ \\ 
			IteR-MRL & 85.98$_{\pm1.62}$        & 1.27$_{\pm0.47}$          & 4.18$_{\pm3.47}$ \\ 
			BS-IRIS & \textbf{87.22}$_{\pm1.98}$ & \textbf{1.12}$_{\pm0.53}$ & \textbf{3.29}$_{\pm3.41}$ \\ 
			\bottomrule[1pt]
	\end{tabular}
	\label{tab:nci}
	
    \begin{tablenotes}
        \footnotesize
        \item[*] No ASSD and HD95 reported.
	 \end{tablenotes}
	\end{threeparttable}
\end{table}

Tables \ref{tab:brats15}, \ref{tab:brats19}, \ref{tab:mmwhs} and \ref{tab:nci} show the quantitative comparison of the five interactive segmentation methods on BraTS2015~\cite{menze:hal-00935640}, BraTS2019~\cite{bakas2018identifying}, MM-WHS~\cite{zhuang2016multi}, and NCI-ISBI2013~\cite{bloch2015nci} respectively. BS-IRIS consistently outperforms all state-of-the-art method for all four datasets, confirming its enhanced robustness. Note that the ``State-of-the-art'' row in each table reports the best-scored automatic segmentation method (without user interactions) for the corresponding challenges. With no bells and whistles compared to the participating networks designed with tricks and hacks, BS-IRIS only use one architecture with the same pre-processing for interactive segmentation, which performs significantly better than the automatic CNNs.

To verify the generalizability of BS-IRIS on same tasks, the best model trained on BraTS2015 is tested on BraTS2019. The model still outperforms the state-of-the-art automatic segmentation method and most of the interactive segmentation methods as shown in Table \ref{tab:brats19}, suggesting the good generalizability on the same tasks.
\begin{figure}
	\centering
	\includegraphics[width=3.3in]{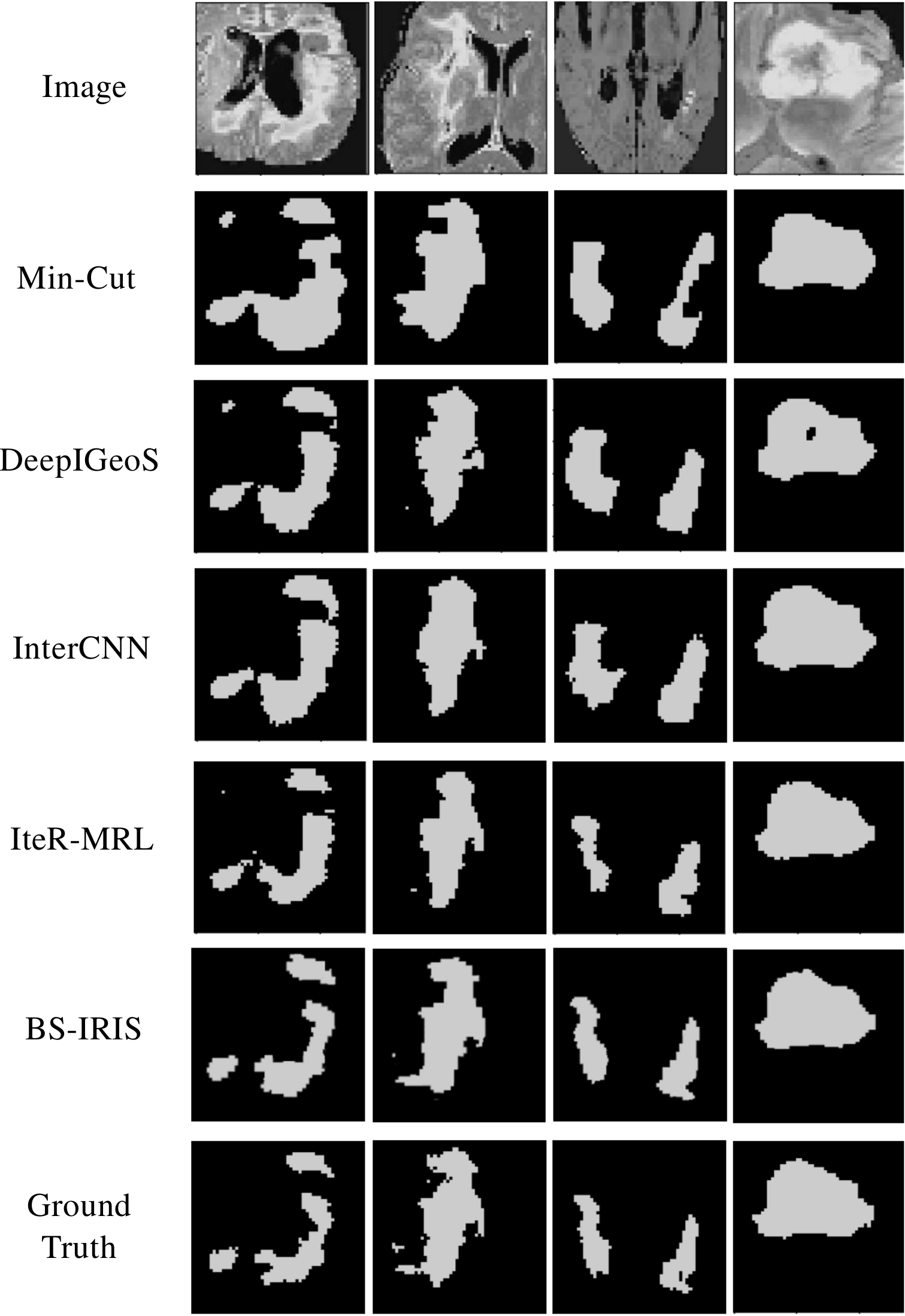}
	\caption{Comparison of segmentation results for interactive methods.}
	\label{update-compare}
\end{figure}

Fig. \ref{update-compare} shows a comparison of the segmentation results for the five interactive methods. With the boundary-aware reward design, BS-IRIS is better at capturing edges and generate more accurate segmentation results than the other methods.

\subsubsection{Improvements in one interaction sequence}
    
\begin{figure}
	\centering
    \includegraphics[width=3.5in]{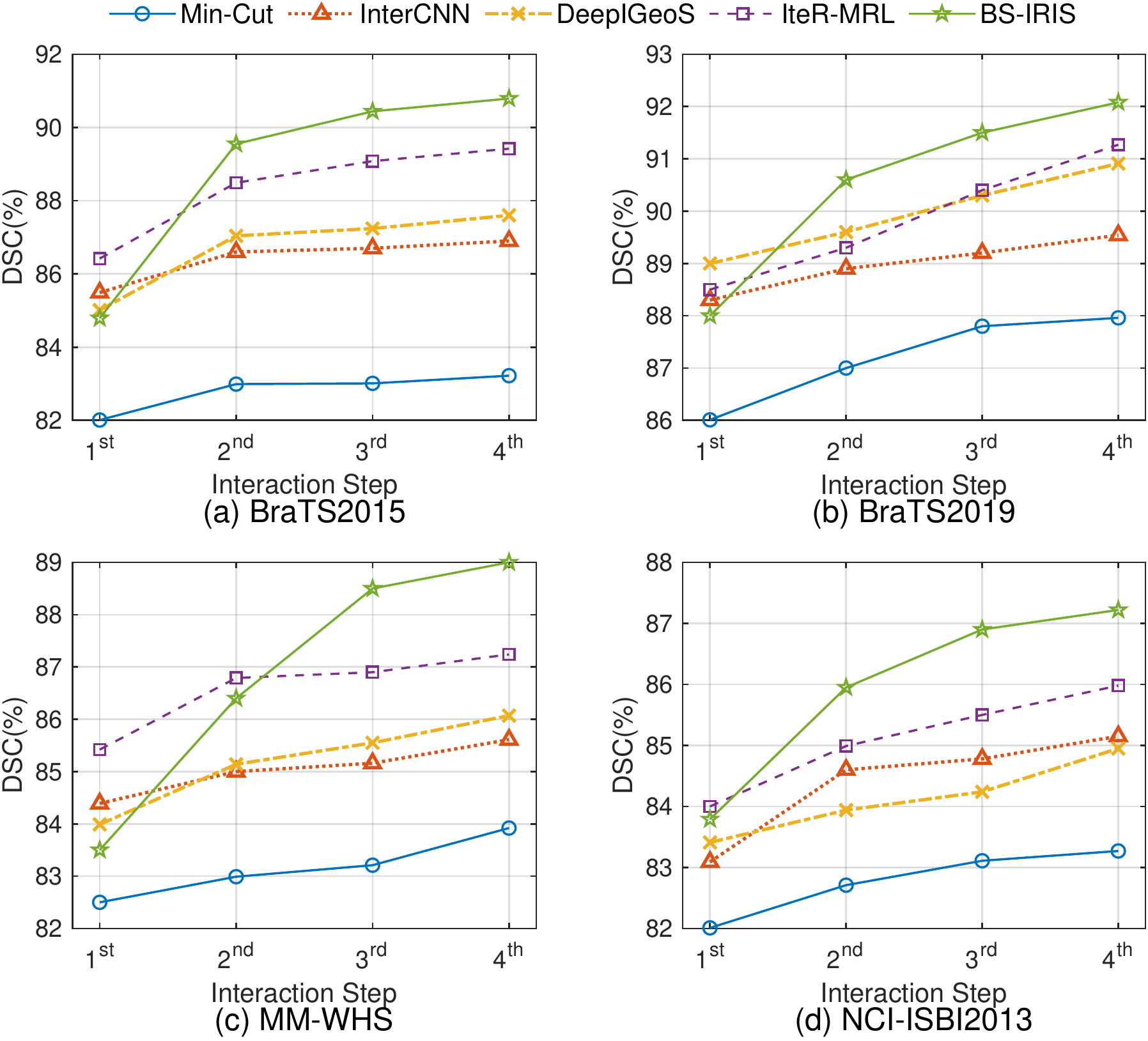}
	\caption{Performance improvements in one interactive  sequence on (a) BraTS2015, (b) BraTS2019, (c) MM-WHS, and (d) NCI-ISBI2013.}
	\label{fig:steps}
\end{figure}

To validate whether BS-IRIS can lead to more rapid improvements than other methods, we also show the performance improvements with interactions during one refinement sequence for four datasets respectively, as shown in Fig. \ref{fig:steps}.
All the interactive methods can improve the performance during refinement iterations. 
However, BS-IRIS has a relatively high and rapid improvement. It starts to outperform other methods in the second iteration, and the advantages are significantly maintained through interactions afterwards.

\subsubsection{Average clicks for achieving specific thresholds}

\begin{table}
    \caption{Comparison of Interactive Segmentation Methods in Required Average Clicks When Achieving Specific Thresholds (DSC) on BraTS2015, BraTS2019, MM-WHS and NCI-ISBI2013. Significant Improvement ($p\text{-value} < 0.05$) Is Marked in Bold}
    \setlength{\tabcolsep}{5pt}
    \begin{tabular}{c|cccc}
    \toprule[1pt]
    Dataset & BraTS2015 & BraTS2019 & MM-WHS & NCI-ISBI2013 \\
    DSC Thr(\%)   & ( $>$ 88)  & ( $>$ 90) & ( $>$ 86)     & ( $>$ 86)   \\ \hline
    Min-Cut      & $>$40    & $>$40    & 35    & $>$40      \\
    InterCNN     & 31        & 27       & 26     & 32        \\
    DeepIGeoS    & 25         & 20      & 23    & 39       \\
    IteR-MRL     & 15         & 16      & 18     & 27     \\
    BS-IRIS      & \textbf{12}         & \textbf{10}      & \textbf{13}     & \textbf{16}    \\ 
    \bottomrule[1pt]
    \end{tabular}
    \label{tab:clicks}
\end{table}

Table \ref{tab:clicks} shows the average clicks required for achieving a specific performance threshold for four datasets. BS-IRIS needs fewer clicks compared with four other interactive methods, which further demonstrates its effectiveness and less labor requirements for the same accuracy.

\subsubsection{Computational time of different methods}

We report the time for the five interactive segmentation methods in Table \ref{tab:time}. With the supervoxel and geodesic transformation computation at each iteration, the BS-IRIS requires much more computation time than the other methods. The computation cost could be lower with some carefully designed implementation. However, it is out of the scope of this paper, and we will leave it for future work. 

\begin{table}
	\centering
    \caption{Comparison of Interactive Segmentation Methods in Computational Time on BraTS2015 with 4 Interactions (6 Clicks/Interaction)}
    \setlength{\tabcolsep}{5pt}
    \begin{tabular}{c |c c c c c }
			\toprule[1pt]
            Method & Min-Cut  & InterCNN & DeepIGeoS & IteR-MRL & BS-IRIS \\
			\hline 
			Sec/Img &  $\approx$ 1.9   & $\approx$ 2 & $\approx$ 2.6  & $\approx$ 2.9  & $\approx$ 4 \\
			\bottomrule[1pt]
    \end{tabular}
    \label{tab:time}
\end{table}

\begin{figure}[htbp]
	\centering
    \includegraphics[width=3.5in]{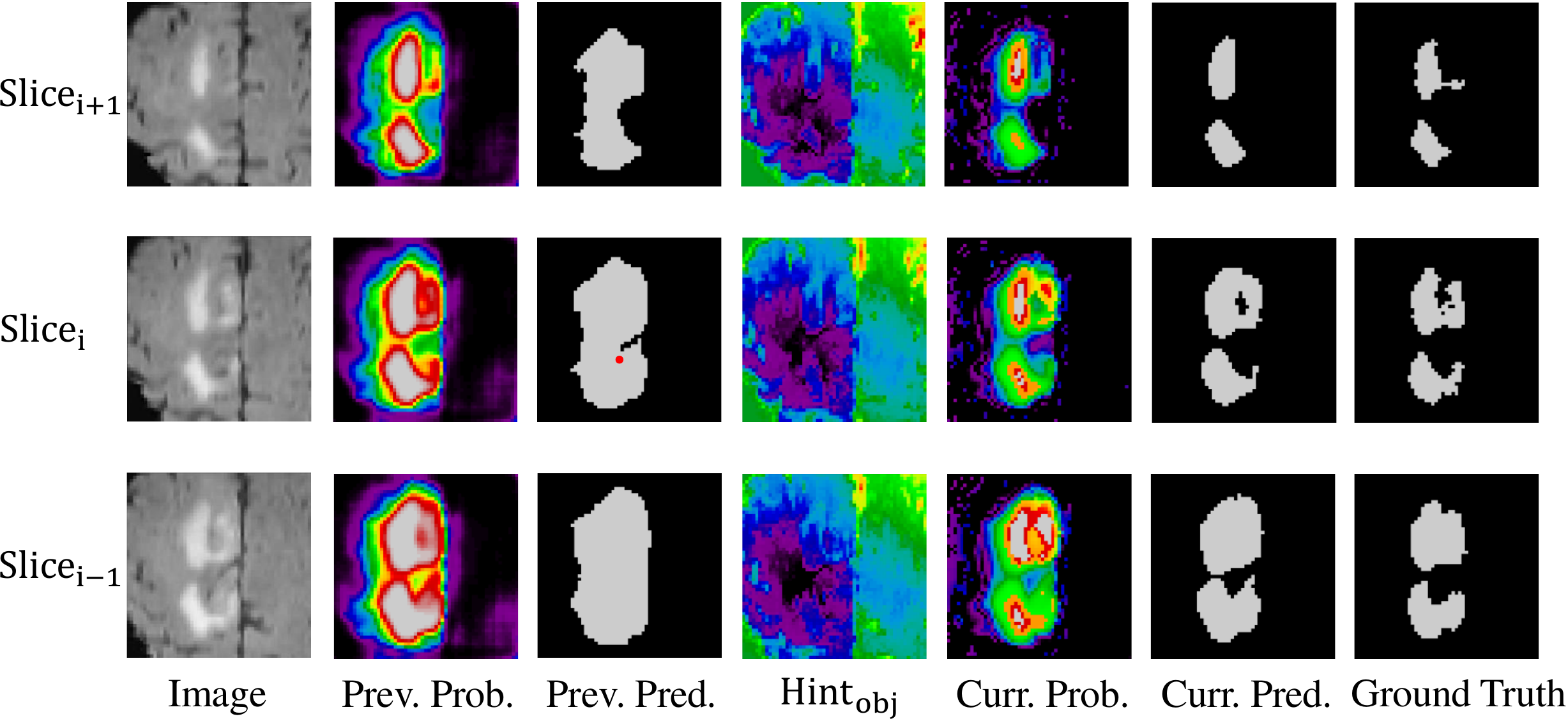}
	\caption{Visualization of one click's influence. The slice with a click and its two neighbor slices are shown. The red point represents user click.}
	\label{update-compare2}
\end{figure}    

\subsection{Visualization of Results}
Fig. \ref{update-compare2} shows the influence of user interaction on probabilities, predictions, and interaction maps. Since the data is 3D, we show the slice with a click (slice$_i$) and its two adjacent slices (slice$_{i-1}$ and slice$_{i+1}$). The black and purple parts on interaction maps are the recommended target area of object regions. We find that the proposed algorithm can successfully correct errors in the local region surrounding the user click. 

\begin{figure}[htbp]
	\centering
	\includegraphics[width=0.43\textwidth]{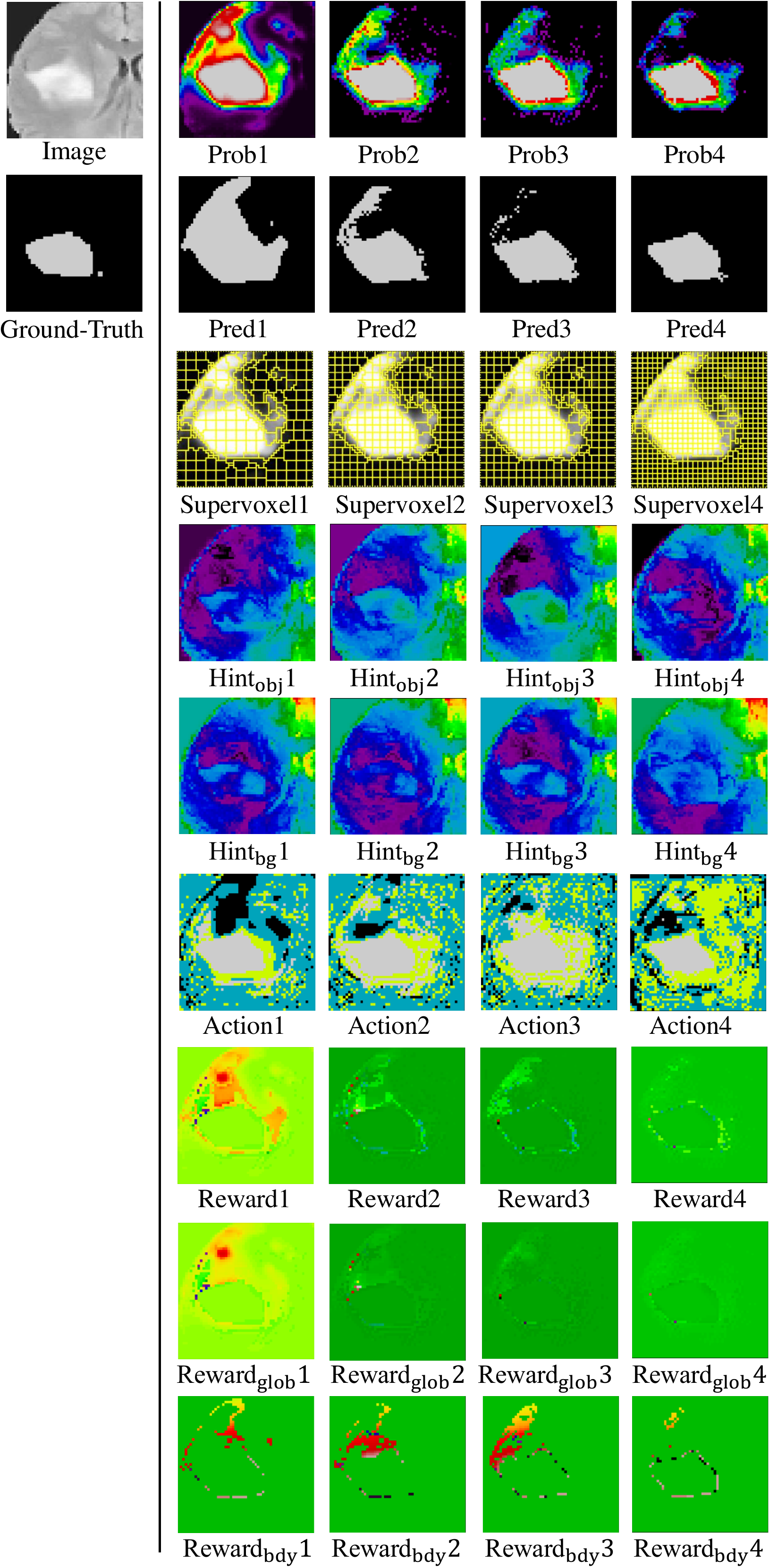}
	\caption{Visualization of segmentation probabilities, predictions, supervoxels, interaction maps, actions and rewards for each iteration. }
	\label{between_step}
\end{figure}

Fig. \ref{between_step} shows the probabilities, predictions, supervoxels, interaction maps, actions, and rewards for $4$ iterations in one interactive sequence. 
One can find that with the interaction maps (Hint$_\text{obj}i$ and Hint$_\text{bg}i$), BS-IRIS succeeds in refining the prediction iteration by iterations. 
For supervoxels masks (Supervoxel$i$), they are gradually dwindling and more fine-grained through iterations, considering the predictions are getting more fine-grained through iteration. 
For the variation of different actions (Action$i$), the number of light green points increased through iterations because this is the smallest value ($\pm0.1$) in the action set, which means agents start to tinily change the probabilities during iterations. 
For the changes of reward, in addition to the boundary-aware reward (Reward$i$), we further show its individual component,  global reward (Reward$_\text{glob}i$) and boundary reward (Reward$_\text{bdy}i$).
Through the iterations, the global reward  becomes less and less obvious. One reason is that the predictions tend to be the close to ground truth, so the relative gain is small. The boundary reward highlights the edge information. The non-green regions shrink by iterations, suggesting the refinement of boundaries.

\section{Conclusion}
In this paper, we propose a novel Boundary-aware Supervoxel-level Iteratively Refined Interactive Segmentation (BS-IRIS) method for 3D medical images using multi-agent reinforcement learning. 
Our method initiatively treats the image segmentation task as an RL problem for each voxel, and explicitly models the dynamic process of interactive segmentation in order to get a rapid improvement at each iteration. 
The experimental results show that it performs better than the automatic and interactive state-of-the-art methods, and it is robust to various datasets. 
For spatially diffused tumors, this supervoxel-clicking method may meet some kinds of mismatch if the tumors is spatially widespread and small.

{\small
\bibliographystyle{ieeetr}
\bibliography{ref}
}

\end{document}